\documentclass{edm_template}

\usepackage[T1]{fontenc}
\usepackage[latin1]{inputenc}
\usepackage{amssymb}
\usepackage{amsmath}
\usepackage{amsfonts}
\usepackage{graphicx}
\usepackage[algo2e,noend,boxed]{algorithm2e}
\usepackage{algorithmic}
\usepackage{algorithm}

\usepackage{xspace}
\usepackage{dsfont}
\usepackage{graphicx}  
\usepackage{verbatim}
\usepackage{appendix}
\usepackage{multirow}
\usepackage{color}
\usepackage{framed}
\usepackage{wrapfig}

\definecolor{shadecolor}{rgb}{.6,.6,.6}
 {\vspace*{0.5\baselineskip}
\begin{center}\begin{minipage}{0.85\textwidth}\begin{shaded}}%
 {\end{shaded}\end{minipage}\end{center} \vspace*{0.5\baselineskip}}

\usepackage[tight,scriptsize]{subfigure}
\newcommand{\denselist}{
\itemsep -2pt\topsep-8pt\partopsep-8pt
}

\numberwithin{equation}{section}

\newcommand{\blockcomment}[1]{ }

\def\reals{\mathbb{R}}

\newbox\subfigbox


\usepackage{booktabs}

\usepackage{hyperref}

\begin{document}

\newcommand{\vecc}{{\mathbf{vec}}}
\newcommand{\sbf}{{\mathbf s}}
\newcommand{\tbf}{{\mathbf{\tau}}}
\newcommand{\bbf}{{\mathbf{b}}}
\newcommand{\Zbf}{{\mathbf Z}}
\newcommand{\ui}{{u_i}}
\newcommand{\vi}{{v_i}}
\newcommand{\unbrs}{{v:v\sim \ui}}
\newcommand{\zvui}{{z^v_{\ui}}}
\newcommand{\zviu}{{z^{\vi}_{u}}}
\newcommand{\tvi}{{\tau_{v_i}}}
\newcommand{\bvi}{{b_{v_i}}}
\newcommand{\vnbrs}{{u:u\sim \vi}}
\newcommand{\Av}{A_v}
\newcommand{\Bv}{B_v}
\newcommand{\Cv}{C_{v'}}
\newcommand{\Dv}{D_{v'}}
\newcommand{\half}{\frac{1}{2}}
\newcommand{\zvu}{z^v_u}
\newcommand{\sui}{s_{u_i}}
\newcommand{\biasvi}{b_{v_i}}
\newcommand{\tauvi}{\tau_{v_i}}
\newcommand{\wbias}{\hat{s}_u^T W \hat{s}_v}
\newcommand{\wbiasu}{\hat{s}_{u_i}^T W \hat{s}_v}
\newcommand{\wbiasv}{\hat{s}_u^T W \hat{s}_{v_i}}
\newcommand{\obspredictu}{\sui + \wbiasu + b_v}
\newcommand{\obspredictv}{s_u + \wbiasv + b_{v_i}}
\newcommand{\obspredict}{s_u + \wbias + b_v}
\newcommand{\svprime}{s_{v'}}
\newcommand{\zuivprime}{z^{u_i}_{v'}}
\newcommand{\bui}{b_{u_i}}
\newcommand{\tauui}{\tau_{u_i}}
\newcommand{\sumovervprime}{\sum_{v':u_i\leadsto v'} }
\newcommand{\obspredictvprime}{s_{v'} + \hat{s}_{v'}^T W \hat{s}_{u_i} + \bui}
\newcommand{\Jbf}{ \mbox {\bf J} }
\newcommand{\Pbf}{ \mbox{\bf P}}
\newcommand{\PGzero}{ $\mbox{\bf PG}_0$\,}
\newcommand{\PGone}{ $\mbox{\bf PG}_1$\,}
\newcommand{\PGonebias}{ $\mbox{\bf PG}_1${\bf-bias}\,}
\newcommand{\PGtwo}{ $\mbox{\bf PG}_2$\,}
\newcommand{\PGthree}{ $\mbox{\bf PG}_3$\,}
\newcommand{\var}{\mbox{var}}

\title{Tuned Models of Peer Assessment in MOOCs\vspace{-3mm}}
%
%
%
%
%

\numberofauthors{6} 
%
\author{
%
%
\alignauthor
Chris Piech\\
       \affaddr{Stanford University}\\
       \email{piech@cs.stanford.edu}
\alignauthor
Jonathan Huang  \\
       \affaddr{Stanford University}\\
       \email{jhuang11@stanford.com}
\alignauthor Zhenghao Chen \\
       \affaddr{Coursera}\\
       \email{zhenghao@coursera.org}
\and  
\alignauthor Chuong Do \\
       \affaddr{Coursera}\\
       \email{cdo@coursera.org}
\alignauthor Andrew Ng \\
       \affaddr{Coursera}\\
       \email{ng@coursera.org}
\alignauthor Daphne Koller \\
       \affaddr{Coursera}\\
       \email{koller@coursera.org}
}
\maketitle \vspace{-4mm}

\begin{abstract}
In massive open online courses (MOOCs), peer grading serves as a critical tool for scaling the grading of complex, open-ended assignments to courses with tens or hundreds of thousands of students. But despite promising initial trials, it does not always deliver accurate results compared to human experts. In this paper, we develop algorithms for estimating and correcting for grader biases and reliabilities, showing significant improvement in peer grading accuracy on real data with 63,199 peer grades from Coursera's HCI course offerings --- the largest peer grading networks analysed to date. We relate grader biases and reliabilities to other student factors such as student engagement, performance as well as commenting style. We also show that our model can lead to more intelligent assignment of graders to gradees.
\end{abstract}
%
\section{Introduction}
The recent increase in popularity of massive open-access online courses (MOOCs),
distributed on platforms such as Udacity, Coursera and EdX, has made it possible
for anyone with an internet connection to enroll in free, university level courses. However while new web technologies  allow for scalable ways to deliver video lecture content,
implement social forums and track student progress in MOOCs,
we remain limited in our ability to evaluate and give
feedback for complex and often open-ended student assignments such as mathematical proofs, design problems and
essays. 
Peer assessment --- which has been historically used 
for logistical, pedagogical, metacognitive, and affective benefits (\cite{sadler06})
 --- 
offers a promising solution that
can scale the grading of complex assignments in courses with tens or even hundreds of thousands of students. 

\begin{figure}[t*]
\begin{center}
\includegraphics[width=.45\textwidth]{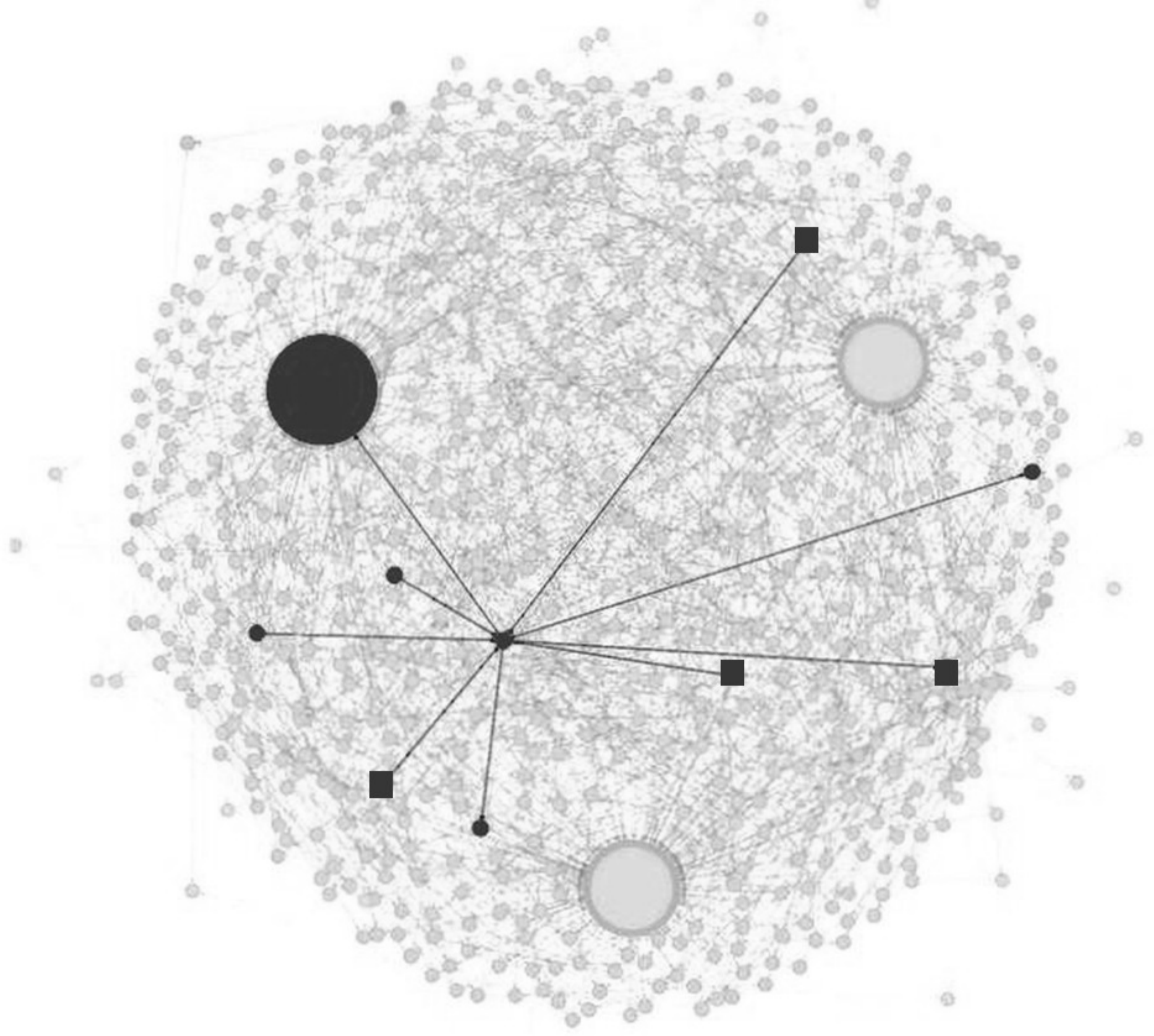}
\end{center}\vspace{-4mm}
\caption{\scriptsize Peer-grading network: Each node is a learner with edges depicting
who graded whom. Node size represents the number of graders for that student.
The highlighted learner shown above graded five students (circular nodes)
and was in turn graded by four students (square nodes). }
\label{fig:gradingnetwork}\vspace{-2mm}
\end{figure}

Initial MOOC-scale peer grading experiments have shown
promise. A recent offering of an online Human Computer Interaction (HCI) course demonstrated that on average, student grades in a MOOC exhibit agreement with staff-given grades \cite{kulkarni13}. Despite their
initial successes, there remains much room for improvement. It was estimated that 43\% of student submissions in the HCI
course were given a grade that fell over 10 percentage points
from a corresponding staff grade, with some submissions up
to 70pp from staff given grades. Thus a critical challenge lies
in how to reliably obtain accurate grades from peers. 

In this paper, we present the largest peer grading networks analysed to date with over $63,000$ peer grades. Our central contribution is to use this unprecedented volume of peer assessment data to extend the discourse on how to create an effective grading system. We formulate and evaluate  intuitive probabilistic peer grading models for estimating submission grades as well as grader biases and reliabilities, allowing ourselves to compensate for grader idiosyncrasies. Our methods improve upon the accuracy of baseline peer grading systems that simply use the median of peer grades by over $30\%$ in 
root mean squared error (RMSE).

In addition to achieving more accurate scoring for peer grading, we also show how fair scores
(where our system arrives at a similar level of confidence about every student's grade)
can be achieved by maintaining estimates of uncertainty of a submission's grade.

Finally we demonstrate that grader related quantities in our statistical model such as bias and reliability
have much to say about other educationally relevant quantities. Specifically we explore summative influences: what variables correspond with a student being a better grader, and formative results: how peer grading affects future course participation. With the large amount of data available to us, we are able to perform detailed analyses of these relationships that would have been difficult to validate with smaller datasets.

Because peer grading is structurally similar in both MOOCs and traditional brick and mortar classrooms, these results shed light on best practices across both mediums. At the same time, our work helps to describe the unique dynamics of peer assessment in a very new setting --- one which may be part of a future with cheaper, more accessible education. 
\vspace{-2mm}
\section{Datasets}\label{sec:datasets}
In this work, we use datasets collected from two 
consecutive Coursera offerings of Human Computer Interaction (HCI), taught by Stanford professor Scott Klemmer.
The HCI courses used a calibrated peer grading system~\cite{russell04} in order to assess weekly student submissions for
assignments which covered a number of different creative design tasks
for building a web site. Calibration required students to correctly assess a training submission before they were allowed
to grade other students' submissions. On every assignment,
each student evaluated five randomly selected submissions
(one of which was a ``ground truth'' submission, discussed below) based on a rubric, and in turn, was evaluated by four
classmates. The final score given to a submission was
determined as the median of the corresponding peer grades.\footnote{
Our description is somewhat of a simplification --- students also performed self-assessments and were given
the higher of the median and their self grade provided that the two were within five percentage points of each other.
We did not consider self assessments in this work.}
Peer grading was anonymized so that students could not see
who they were evaluating, or who their evaluators were.
See Kulkarni et al.~\cite{kulkarni13} for details of the peer grading system.

After the first offering (HCI1), the peer grading system was refined in several ways. Among other things, HCI2 featured a modified rubric that
addressed some of the shortcomings of the original peer grading scheme. Additionally, peer graders were divided into language groups
(English and Spanish) to address concerns of being graded
by a non-native speaker as well as the observed ``patriotic
grading effect''~\cite{kulkarni13}.  
Counting just those who submitted at least one assignment 
in the English offerings of the class, there were 3,607 students 
from the first offering (HCI 1) and 3,633 students
from the second offering (HCI 2).  These students came from diverse backgrounds
 (with a majority of students from outside of the United States). 
 Collectively, these 7,240 students from around the world 
created 13,972 submissions, receiving 63,199 peer grades in total. See Table~\ref{tab:datasets} for a summary of the dataset.
In our work, we used the data
from HCI2 as a hold out set. We formulated our models based only on exploratory experiments performed using the HCI1 dataset, testing on the second HCI
class only after having finalized our theories about which
models were useful.

\begin{table}[t!]
\caption{Data Sets}
\begin{center}
\begin{tabular}{rcc}
\hline
 &  First HCI & Second HCI \\
\hline
Students  & 3,607  &  3,633\\
Assignments & 5  &  5\\
Submissions & 6,702 & 7,270\\
Peer Grades & 31,067  &  32,132\\
\hline
\end{tabular}
\end{center}
\label{tab:datasets}
\vspace{-4mm}
\end{table}

The software for the peer grading framework
used by the HCI courses was designed to accommodate experimental validation of peer grading. A small number (3-5)
of submissions for each assignment were marked as ``ground
truth'' and were then graded by the course staff. Since there
were only a few ground truth submissions and each student
graded at least one per week, the ground truth submissions
were ``super-graded'' and had, on average, 160 assessments.
Of note, the students were not told that one of the submissions they were assigned to mark belonged to the ground truth set. For
example, Figure~\ref{fig:gradingnetwork} shows the network of gradee-grader relationships on Assignment 5 of HCI1, where the three super-graded ground truth submissions are clearly visible. 
\section{Probabilistic models of peer \\grading in MOOCs}
The ideal peer grading system for a MOOC should satisfy the following desiderata: it should (1) provide
highly reliable/accurate assessment, (2) allocate a balanced
and limited workload across students and course staff, (3)
be scalable to class sizes of tens or hundreds of thousands
of students, and (4) apply broadly to a diverse collection of
problem settings. 
In this section we discuss a number of ways to formulate a
probabilistic model of peer grading to address these desiderata. The models 
that we introduce allow for
us to algorithmically compensate for factors such as grader
biases and reliabilities while maintaining estimates of uncertainty in a principled way.

Through our paper, we will use the following notation.
We refer to the collection of all submissions to a homework assignment as $U$, and specific submissions
indexed as $u\in U$.  We assume in this paper that each student corresponds to a unique homework submission
per assignment, and thus refer to students (users) and submissions interchangeably.
The collection of all graders is denoted by $G$, and specific graders
by $v\in G$.   Note that graders are themselves students with submissions.
Finally, we use the notation $v\rightarrow u$ to mean that grader $v$ grades submission $u$.  For example,
the set $\{u\,:\,v\rightarrow u\}$ refers to the collection of submissions graded by a single student $v$.

Our models assume the existence of the following quantities which are either observed or 
latent (unobserved) variables which we wish to estimate.\vspace{-3mm}
\begin{itemize}\denselist
\item {\bf True scores:}
We assume that every submission $u$ is associated with a \emph{true underlying
score}, denoted $s_u$, which is unobserved and to be estimated. 
\item {\bf Grader biases:}
Every grader $v$ is associated with a bias, $b_v\in \reals$.  These bias
variables reflect a grader's tendency to either inflate or deflate her assessment by a certain
number of percentage points.  
\item {\bf Grader reliabilities:}
We also model grader reliability, $\tau_v \in \reals^+$, reflecting how close on average
a grader's peer assessments tend to land near the corresponding submission's true score after having
corrected for bias.  In the models
below, $\tau_v$ will always refer to the \emph{precision}, or inverse variance of a normal distribution.
\item {\bf Observed grades:} 
Finally, $z^v_u\in \reals$ is the observable score given by grader $v$ to submission $u$.
The collection of all observed peer grades is denoted as  $Z=\{z^v_u\}$.
\end{itemize}
\begin{figure*}
\begin{center}
\subfigure[]{
\includegraphics[width=.30\textwidth]{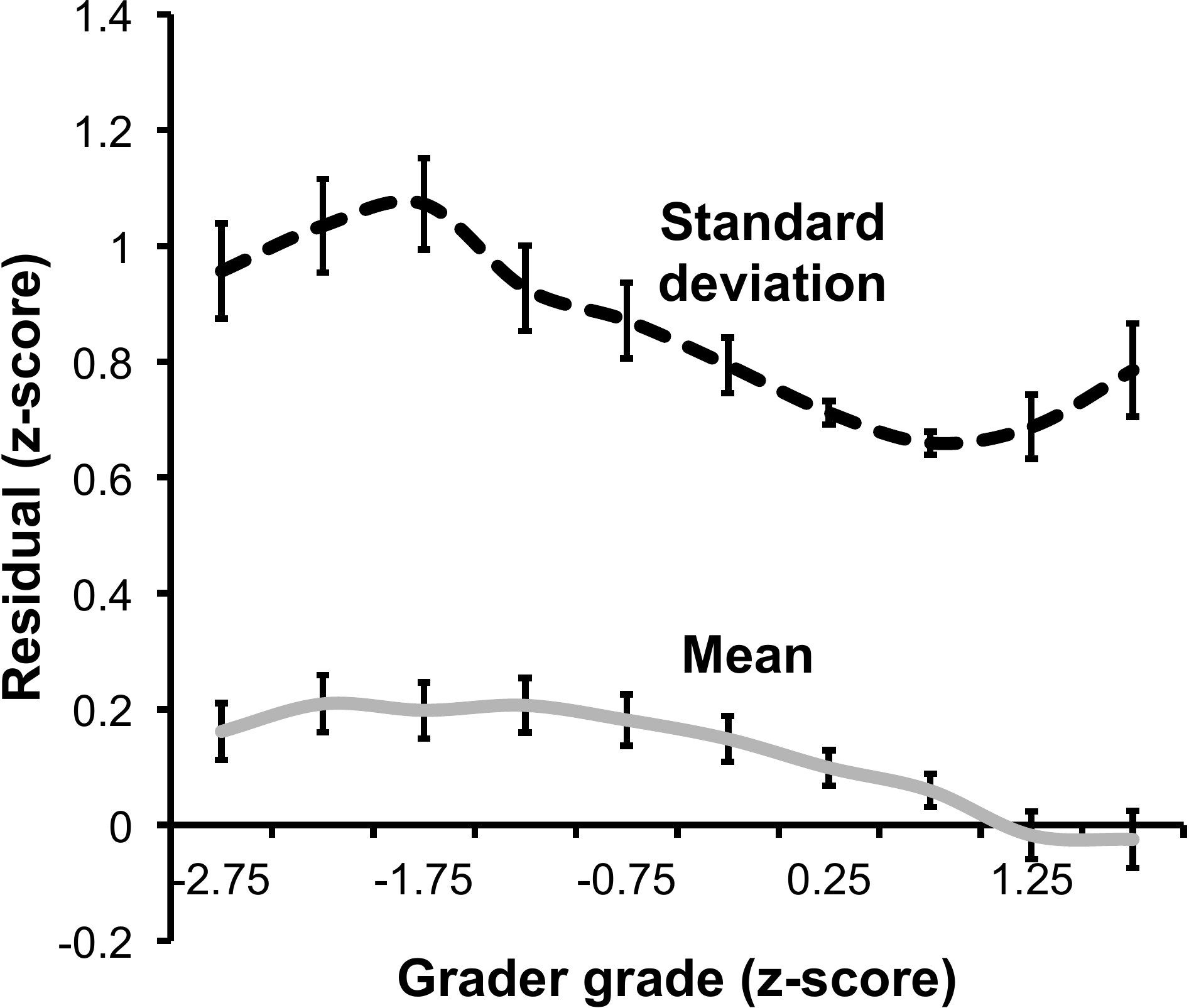}
\label{fig:gradergrade_v_residual}
}\;\;
\subfigure[]{
\includegraphics[width=.30\textwidth]{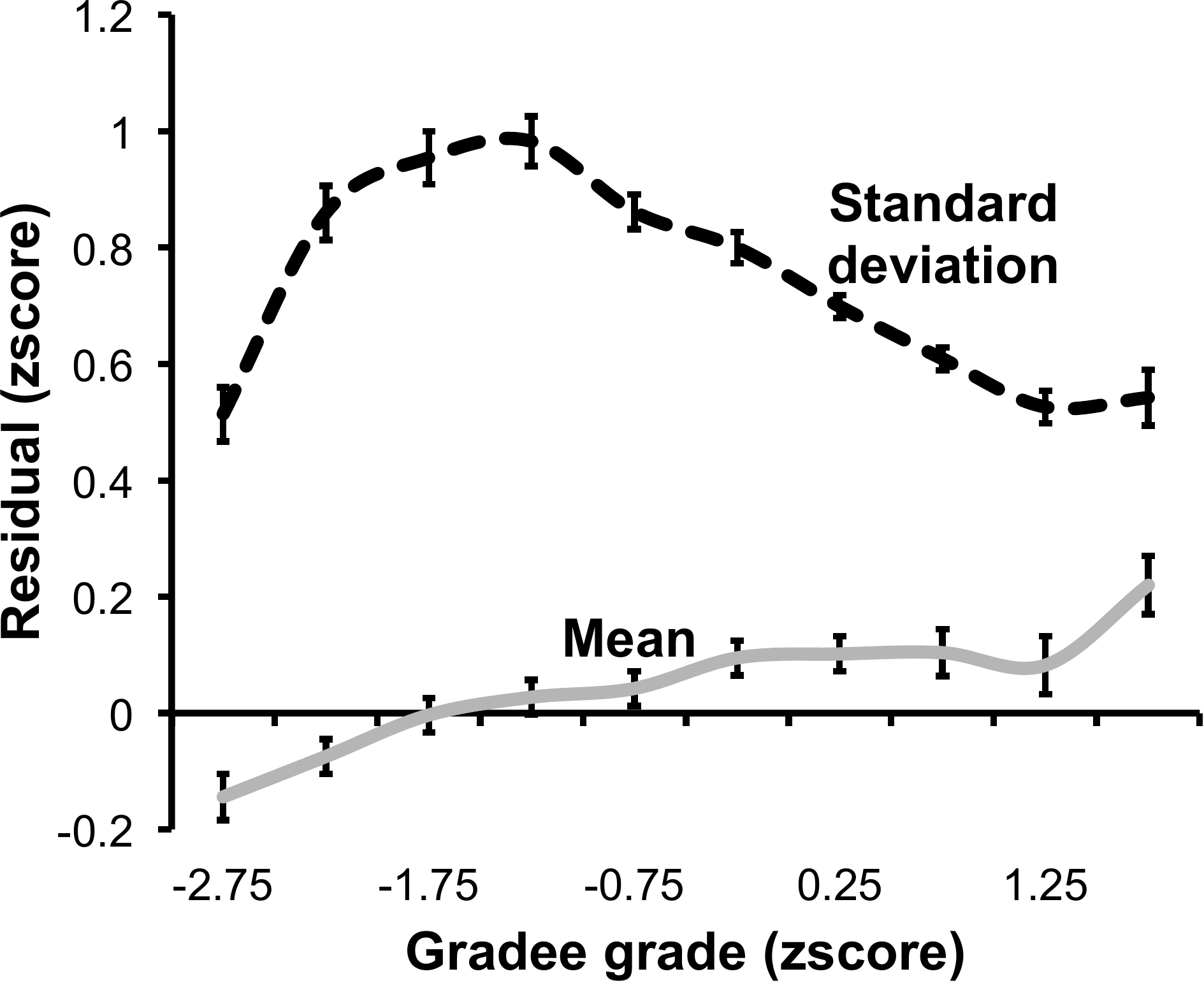}
\label{fig:gradeegrade_v_residual}
}\;\;
\subfigure[]{
\includegraphics[width=.30\textwidth]{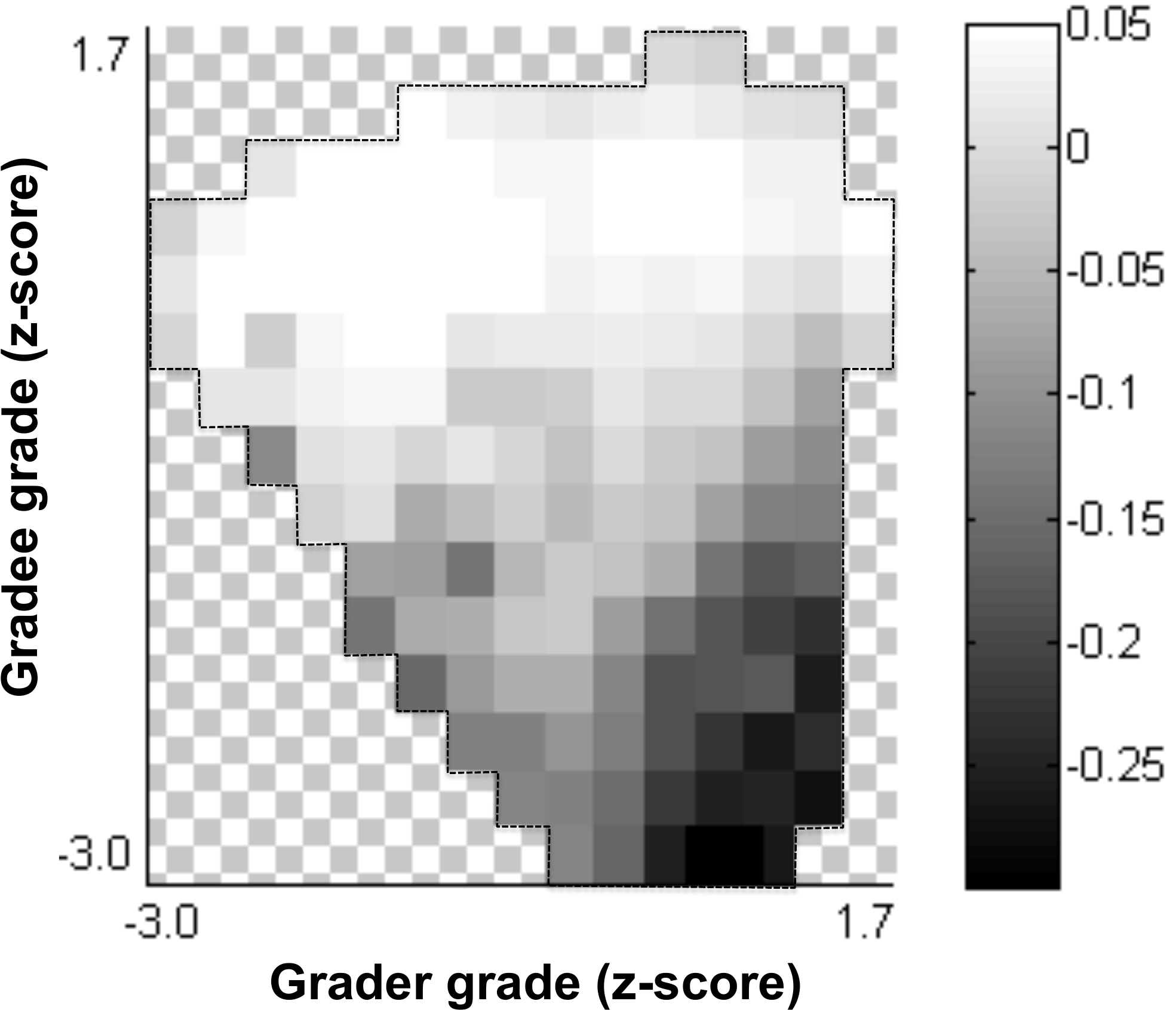}
\label{fig:gradergradee}
}
\end{center}\vspace{-7mm}
\caption{\scriptsize 
\subref{fig:gradergrade_v_residual} The relationship between a grader's homework performance (her grade)
and statistics (mean/standard deviation) of grading performance (residual from true grade).
\subref{fig:gradeegrade_v_residual} The relationship between a gradee's homework performance against statistics of assessments
for her submissions.
\subref{fig:gradergradee} Visualization of all three variables simultaneously, where intensity reflects
the mean residual $z$-score.  Empty boxes mean that there is not enough data available to compute a reliable estimate.
}
\label{fig:graders}\vspace{-3mm}
\end{figure*}

\subsection{Models}
Below we present, in order of increasing complexity, three statistical models that we have found to be particularly effective.
\vspace{-6mm}
\paragraph{Model \PGone (Grader bias and reliability)}
Our first model, \PGone puts prior distributions over the latent variables and
assumes for example that while an individual grader's bias may be nonzero,
the average bias of many graders is zero.
Specifically, \vspace{-7mm}

{\footnotesize
\begin{align*}
\mbox{(Reliability)}\; \tau_v &\sim  \mathcal{G}(\alpha_0,\,\beta_0) \;\mbox{for every grader $v$},\\
\mbox{(Bias)}\;b_v &\sim \mathcal{N}(0,\,1/\eta_0) \;\mbox{for every grader $v$},\\
\mbox{(True score)}\;s_u&\sim \mathcal{N}(\mu_0,\,1/\gamma_0)  \;\mbox{for every user $u$, and}\\
\mbox{(Observed score)}\;z^v_u &\sim \mathcal{N}( s_u + b_v,\, 1/\tau_v ), \\
	&\qquad \;\mbox{for every observed peer grade.} 
\end{align*}\vspace{-8mm}
}

$\mathcal{G}$ refers to a gamma distribution with fixed hyperparameters $\alpha_0$, $\beta_0$, while
$\eta_0$, and $\gamma_0$ are hyperparameters for the priors over biases and true scores, respectively.
In our experiments, we also consider a simplified version of
Model \PGone in which the reliability of every grader is fixed
to be the same value. We refer to this simpler model in which
only the grader biases are allowed to vary as \PGonebias. 

\vspace{-6mm}
\paragraph{Model \PGtwo (Temporal coherence)}
The priors for reliability and bias can play a particularly important role in the above model
due to the fact that we typically only have about 4-5 grades to estimate the bias and reliability
of each grader.  A simple way to obtain more data per grader 
is to leverage observations made about the grader from previous assignments.  To pose a model, we must understand the relationship
of a grader's bias and reliability at homework $T$ to that at homework $T'$.  Is it the same or does it change over time?

To answer this question, we examine the correlation between the estimated biases from Model \PGone using 
the HCI1 dataset (see Section~\ref{sec:datasets}). Between consecutive assignments, 
a grader's biases have a Pearson correlation of 0.33 which
represents a utilizable consistency. Grader reliability, on the other hand, has a low correlation.
We therefore posit Model \PGtwo which allows for grader biases
at homework $T$ to depend on those at homework $T-1$ (and implicitly, on all prior homeworks).   Specifically, Model \PGtwo assumes:
\vspace{-5mm}

{\footnotesize\allowdisplaybreaks
\begin{align*}
\tau_v^{(T)} &\sim  \mathcal{G}(\alpha_0,\beta_0) \;\mbox{for every grader $v$},\\
b_v^{(T)} &\sim \mathcal{N}(b_v^{(T-1)},1/\omega_0) \;\mbox{for every grader $v$},\\
s_u^{(T)}&\sim \mathcal{N}(\mu_0,1/\gamma_0) \; \mbox{for every user $u$, and}\\
z^{v,(T)}_u &\sim \mathcal{N}( s_u^{(T)} + b_v^{(T)}, 1/\tau_v^{(T)} ), \\
	&\qquad \;\mbox{for every observed peer grade.} 
\end{align*}\vspace{-8mm}
}

Model \PGtwo requires that we normalize grades across different homework assignments to a consistent scale. In our
experiments, for example, we have noticed that the set of grader
biases had different variances on different homework assignments. Using
a normalized score ($z$-score), however, allows us to propagate a student's underlying bias while remaining robust to assignment
artifacts.

Note that while a model which captures the dynamics of true scores and reliabilities across assignments can be similarly imagined,
we have focused only on the dynamics of bias for this work (which 
contributes the most towards improved accuracy while still being equitable). 
\vspace{-5mm}
\paragraph{Model \PGthree (Coupled grader score and reliability)}
A unique aspect of peer grading is that graders are themselves
students with submissions being graded. 
Consequently, it is of interest to understand and model the relationship between
one's grade and one's grading ability --- 
for example, knowing that a student
scored well on his assignment may be cause for placing more
trust in that student as a grader, and vice versa. 

In Figure~\ref{fig:graders}, we show
experiments exploring the relationships between the grader
specific latent variables. In particular, we observe
that high scoring students tend to be somewhat more reliable as graders (see details of the experiment in 
Section~\ref{sec:experiments}). Model \PGthree formalizes this
intuition by allowing the reliability of a grader to depend on
her own grade, and assumes the following:\vspace{-5mm}

{\footnotesize\allowdisplaybreaks
\begin{align*}
b_v &\sim \mathcal{N}(0,\,1/\eta_0) \;\mbox{for every grader $v$},\\
s_u&\sim \mathcal{N}(\mu_0,\,1/\gamma_0)  \;\mbox{for every user $u$, and}\\
z^v_u &\sim \mathcal{N}\left( s_u + b_v,\, \frac{1}{ \theta_1 s_v+ \theta_0 } \right),\\
	&\qquad  \;\mbox{for every observed peer grade.} 
\end{align*}\vspace{-6mm}
}

Note that Model \PGthree extends\PGone by introducing new dependencies, allowing us to use a student's submission
score to estimate her grading ability.
At the same time Model \PGthree is more constrained, forcing grader reliability to depend on a single parameter
instead of being allowed to vary arbitrarily, and thus prevents our model from overfitting.
\vspace{-3mm}
\paragraph{Ethics and Incentives} 
If we are to use probabilistic inference to score students in a MOOC,
the end goal could not simply be to optimize for accuracy.
We must also consider fairness when it comes to deciding
what variables to include in the model. It might be tempting, for 
example, to include variables such as race, ethnicity and gender into a model for better accuracy, but almost everyone
would agree that these factors could not be fairly used within a
scoring mechanism even if they improved prediction accuracy. Another 
example might be to model the temporal coherence of student grades (we observe a particularly strong
temporal correlation between students' grades --- with 0.46 Pearson
coefficient --- of consecutive homework assignments). But incorporating this temporal coherence for students scores
into a scoring mechanism would
not allow for students to be given a ``clean slate'' on each
homework.

Model \PGthree allows for the inferred true score of a submission
to depend on graders' scores, which may seem contentious,
but the dependence is weak, only affecting the influence by
a particular grader on the final prediction, which is desirable. Interestingly, using the more complex scoring mechanism from Model \PGthree may 
in fact incentivize for good
grading. In particular, a student's grade is influenced by
how closely her assessments as a grader match those of other
graders who graded the same assignments. Consequently, by
allowing for student grades to depend on their performance
as graders, Model \PGthree used as a scoring mechanism may
incentive students to put more effort into grading.
\vspace{-3mm}
\subsection{Inference and evaluation.}
Given a probabilistic model of peer grading such as those discussed above, 
we would like to infer the values of the unobserved variables such as the  true score of 
every submission, or the bias and reliability of each student as a grader.
Inference can be framed as the problem of computing the posterior distribution over the 
latent variables conditioned on all observed peer grades (e.g., $P(\{s_u\}_{u\in U}, \{b_v\}_{v\in G}, \{\tau_v\}_{v\in G}  \;|\; Z)$).

Computing this posterior is nontrivial, since all of the variables are correlated 
with each other.  For example, having good estimates of the biases of all of the graders to submission $u$ ($\{b_v: v\rightarrow u\}$)
would allow us to better estimate $u$'s true score, $s_u$.  However to estimate each bias $b_v$, we would have to have good estimates of the true scores of \emph{all} of the submissions graded
by $v$ ($\{s_u:v \rightarrow u\}$).
We must therefore reason circularly, in that --- if we knew
every submission's true scores, we would be able to easily compute posterior distributions
over grader biases (and reliabilities), but in order to estimate these biases, we must know
the true score of each submission.  

To address this apparent chicken and egg problem, we turn to simple approximate inference methods. 
In the experiments reported in Section~\ref{sec:experiments}, we use Gibbs sampling~\cite{geman84}, which produces
a collection of samples from the (approximate) desired posterior distribution. These samples can then be used to estimate various 
quantities of interest. For example, given
samples $s^1_u, s^2_u,\dots, s^T_u$ from the posterior distribution over 
the true score of submission $u$, we estimate the true score as: $\hat{s}_u \equiv \frac{1}{T} \sum_{t=1}^T s^t_u$. 
We can also use the samples to quantify the uncertainty of our prediction by estimating the variance of the samples from the
posterior, which we use in Section~\ref{sec:experiments} when we examine peer
grading efficiency. Note that while the ordinary Gibbs sampling algorithm can be performed in ``closed form'' for Models
\PGonebias, \PGone and \PGtwo, Model \PGthree requires numerical approximation due to the coupling of a submission's true
score $s_u$ with that of its grader, $s_v$. 
We discuss details in the Appendix.\footnote{See accompanying appendix at \url{www.stanford.edu/~cpiech/bio/papers/appendices/edm13_appendix.pdf}}  
Visually we observe rapid mixing for our Gibbs chains, and in the experiments shown in Section~\ref{sec:experiments}, we use
800 iterations of Gibbs sampling, discarding the initial 80 burn-in samples.


Expectation-maximization (EM) is alternative approximate inference approach, where we treat the true scores and grader biases as parameters
and then use an iterative coordinate descent based algorithm to obtain point estimates of parameters.  
In practice, we find that both the Gibbs and EM approaches behave
similarly.  In general EM has the advantage of being significantly faster while obtaining posterior credible intervals is more natural using Gibbs. On the peer grading dataset the two methods produce analogous results. For example, \PGone with Gibbs and EM have RMSE scores of 5.42 and 5.43 on the first dataset respectively and with Gibbs running in roughly 5 minutes and EM running in 7 seconds.
We refer the reader to the appendix for the full algorithmic details of Gibbs as well as EM.

\newcommand{\ra}[1]{\renewcommand{\arraystretch}{#1}}
\begin{table*}
\caption{Comparison of models on the two HCI courses}
\centering\footnotesize
\ra{1.3}
\begin{tabular}{@{}rcccccccccccc@{}}
\toprule[1.5pt]
& \multicolumn{5}{c}{HCI 1} & \phantom{abc}& \multicolumn{5}{c}{HCI 2} \\
\cmidrule{2-6} \cmidrule{8-12} 
& Baseline & \PGonebias & \PGone & \PGtwo & \PGthree &&  					Baseline & \PGonebias & \PGone & \PGtwo & \PGthree  \\ \midrule
RMSE & 7.95 & 5.42 & 5.40 & 5.40 & {\bf 5.30} 			&& 6.43 & 4.84 & 4.81 & 4.75 & {\bf 4.73} \\
\% Within 5pp &  51& 69& 69& {\bf 71} & 70			&& 59& 72& 73& 73 & {\bf 74} \\
\% Within 10pp & 81& 92& 94& 94& {\bf 95}			&& 88& 96& 96& {\bf 97} & {\bf 97} \\
Mean Std & 7.23& 5.00& 4.96& 4.92& {\bf 4.77}		&& 6.19& 4.57& {\bf 4.52} & 4.53 & {\bf 4.52} \\
Worst Grade & -43& -34& {\bf -30} & -32& {\bf -30}		&& -36& -26& -26& {\bf -25} & -26\\
\bottomrule[1.5pt]
\end{tabular}
\label{tab:results}\vspace{-4mm}
\end{table*}
\vspace{-3mm}
\paragraph{Evaluation}
To measure peer grading accuracy, we repeatedly simulate
what score would have been assigned to each ground truth
submission had it been peer graded. Our evaluation of how
well we would have graded a single ground truth submission
uses a two step methodology (based on the evaluation method of~\cite{kulkarni13}): (1) We run inference using
all of our data, except the peer grades of the ground truth
submission being evaluated. This gives us an estimate of each grader's biases and reliabilities as well as model
priors that were independent of the submission being evaluated. 
(2) We run simulations where we sampled four student
assessments randomly from the pool of peer grades for the
ground truth submission, estimate the submission's
grade using the sample of assessments and recorde the
residual between our estimated grade and the ``true'' grade.
For each ground truth submission we run 3000 such simulations, from which we report the RMSE, the number of simulations which fell
within five, and ten percentage points of the true score, the
average standard deviation of the errors over each ground
truth and the worst misgrade that the simulations produced.

An interesting issue is whether one should consider the ``true''
grade of a ground truth submission to be the score given by
the staff, or the consensus from the hundreds of students
that assessed the submission. For our datasets, we believe
that the discrepancy between staff grade and student consensus typically results from ambiguities in the rubric and
elect to use the mean of the student consensus on a ground
truth submission as the true grade. One interesting observation that came from our exploration: peer graders in our
datasets have a tendency to grade towards the mean, inflating
grades for low-scoring submissions and deflating grades for
high-scoring submissions. 
 We remark that
while our experiments were run in an ``unsupervised'' fashion, it would be reasonable to use staff grades
in the training process in order to encourage the model to
place more trust in students who consistently grade like the
instructors.

We compare each of our probabilistic models to the grade
estimation algorithm used on Coursera's platform. In the
baseline model, the score given to students is the median of
the four peer grades they received. Specifically, the baseline
estimation does not take into account individual grader's
biases and reliabilities. Nor does it incorporate prior knowledge about the distribution of true grades. 
\vspace{-3mm}
\section{Experimental results}\label{sec:experiments}
\begin{figure*}
\begin{center}
\subfigure[]
{
\includegraphics[width=.23\textwidth]{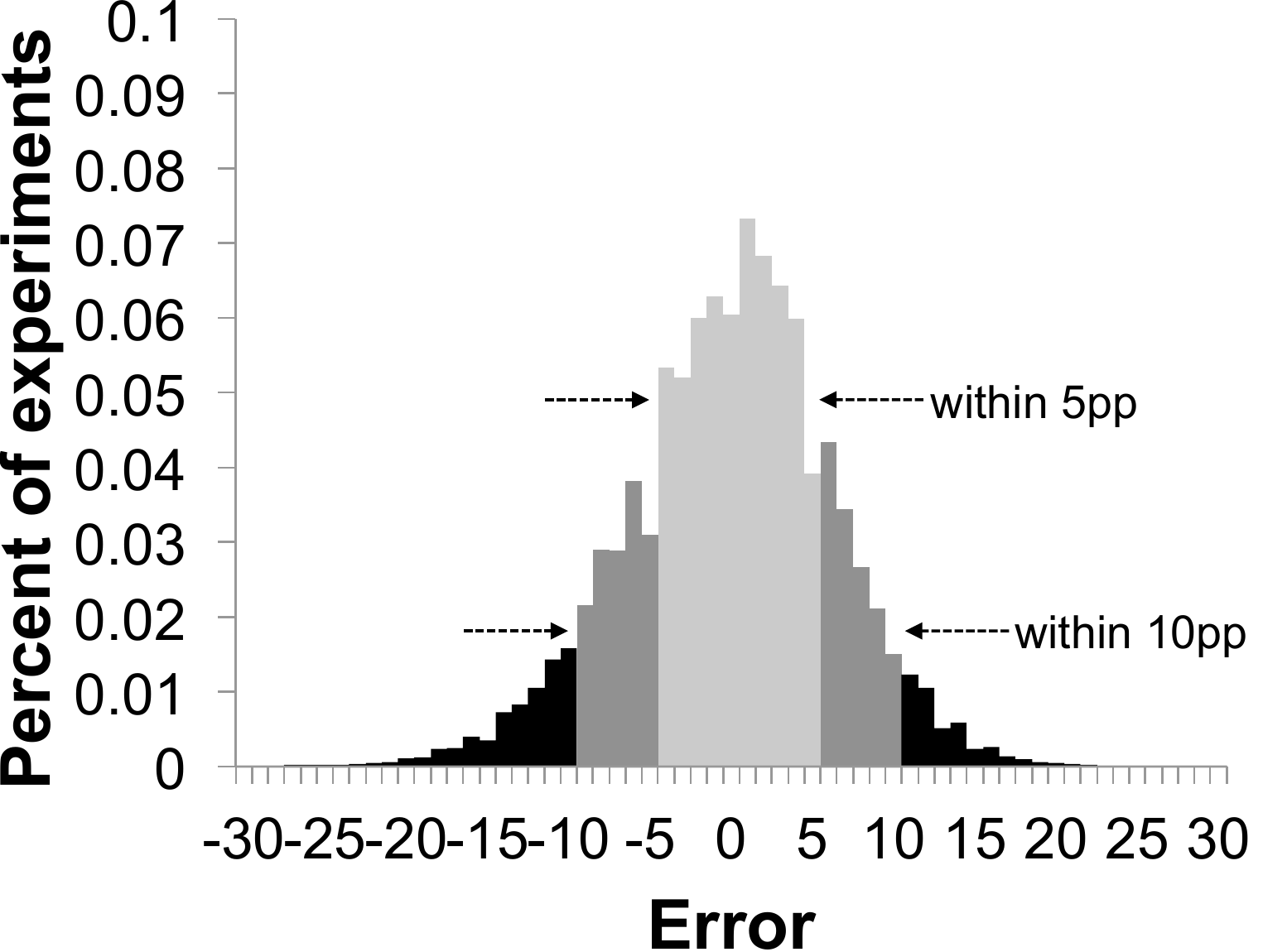}
\label{fig:baseline}
}
\subfigure[]
{
\includegraphics[width=.23\textwidth]{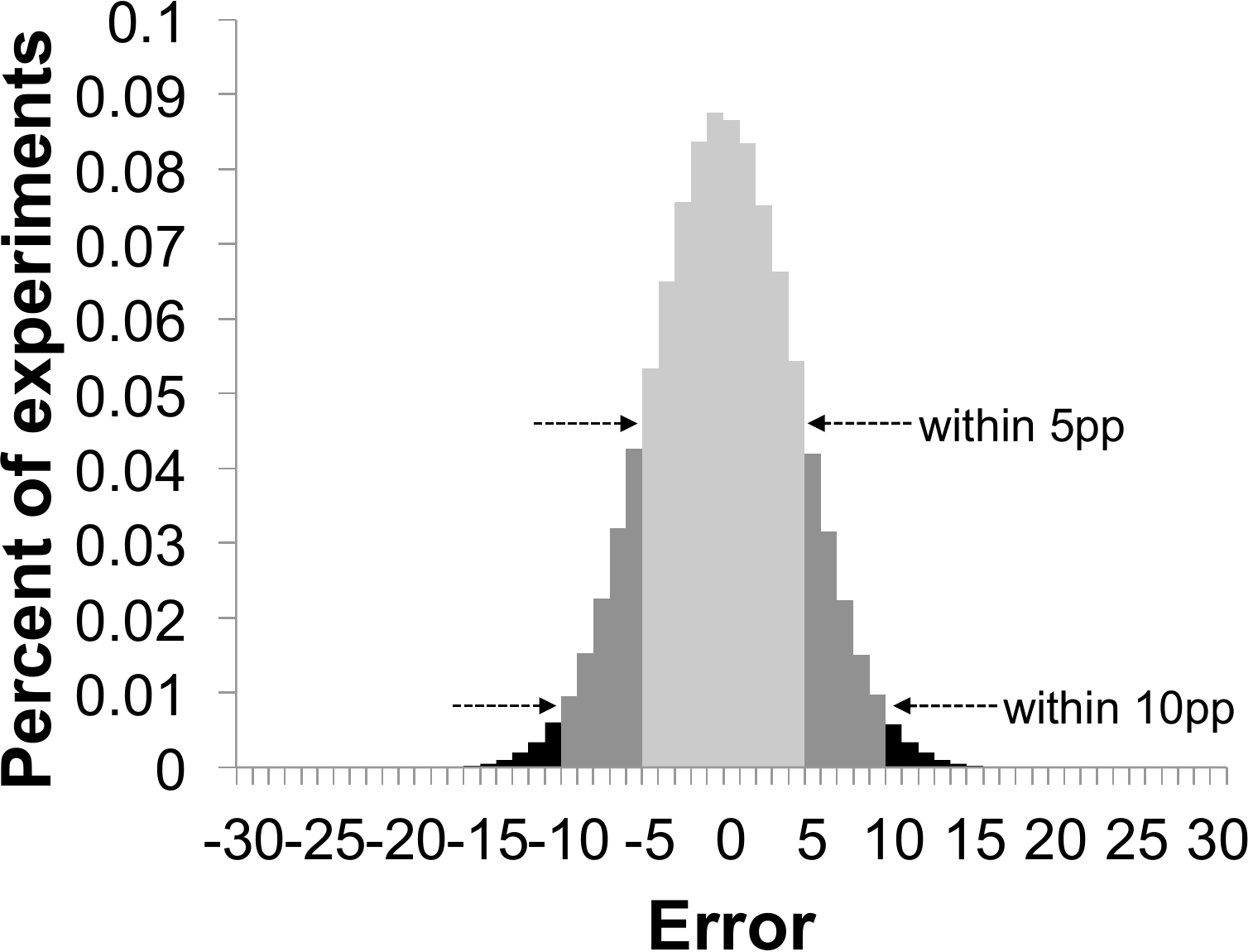}
\label{fig:improved}
}
\subfigure[]
{
\includegraphics[width=.22\textwidth]{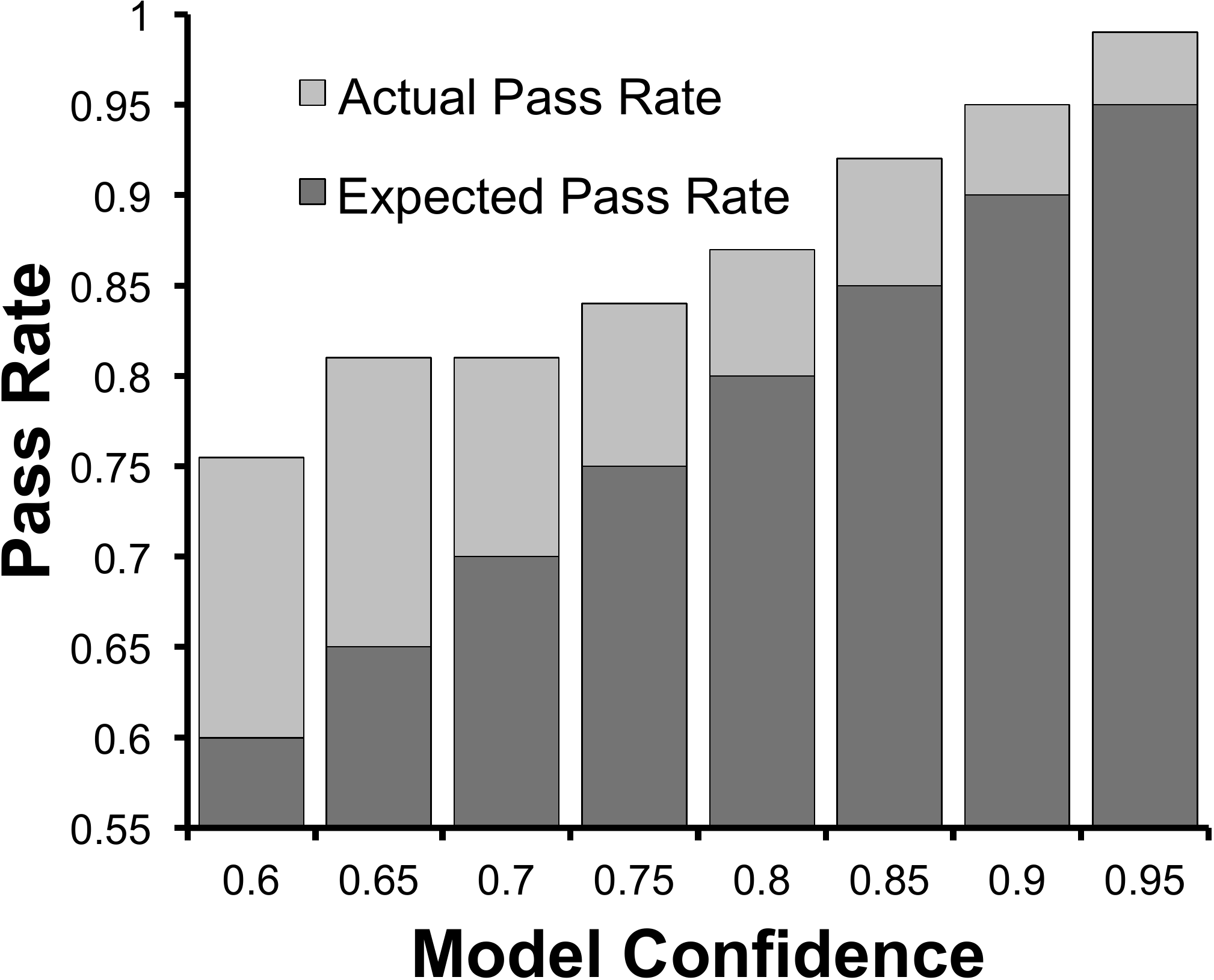}
\label{fig:confidence}
}
\subfigure[]
{
\includegraphics[width=.20\textwidth]{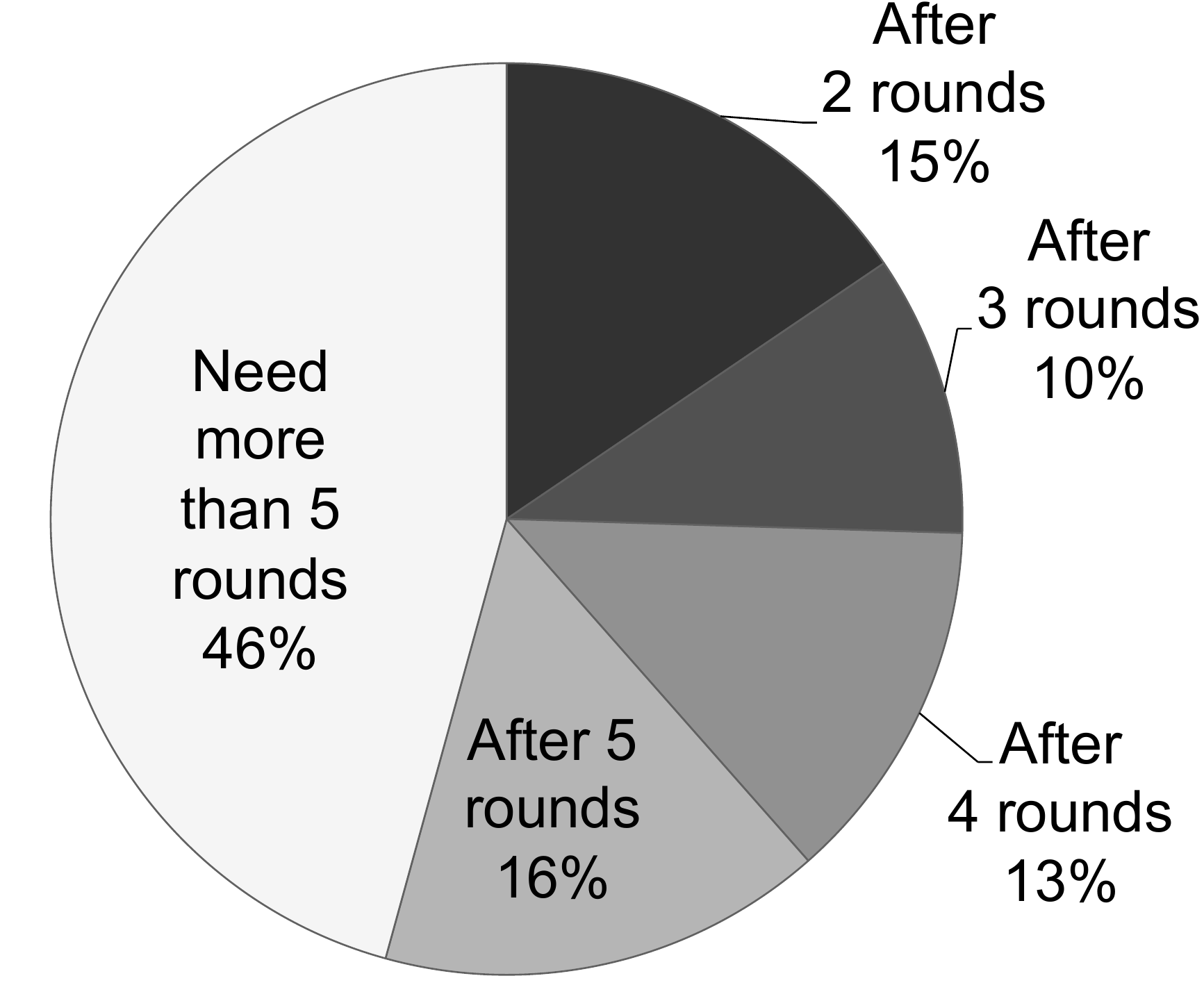}
\label{fig:piechart}
}
\end{center}\vspace{-7mm}
\caption{\scriptsize
\subref{fig:baseline} Histogram of errors made using the baseline (median) scoring mechanism. 
 \subref{fig:improved} Histogram of errors using \PGthree.  
 \subref{fig:confidence}  A comparison of model confidence ($x$-axis) and actual success rate of predictions ($y$-axis), where
 being above the diagonal (dark bars) is better.
 \subref{fig:piechart} Number of submissions for which our model can declare ``confidence''
 after $K$ rounds of grading.
 }
\label{tab:predacc}
\vspace{-4mm}
\end{figure*}
\vspace{-1mm}
\subsection{Accuracy of reweighted peer grading}
Using probabilistic models leads to substantially higher grading accuracy. In our experiments we are able to reduce the
RMS error on our prediction of the ground truth grade by
33\% from 7.95 to 5.30. Similarly, on the second offering of
the course we were able to reduce error by 31\% from 6.43
to 4.73. For the second offering, this means that the number of students who received grades within 10 percentage
points (pp) of their grade increased from 88\% to 97\%. Figures~\ref{fig:baseline},~\ref{fig:improved} show the effect of using Model~\PGthree
as a scoring mechanism on the histogram of grading errors and Table~\ref{tab:results} shows the
complete results for each model. Due to course improvements, we observe that students
in HCI2 were significantly more consistent as graders compared to students in HCI1. However, we remark that every
one of our models run on HCI1 outperforms the baseline
grading system run on HCI2 with respect to every metric,
indicating that the best gains in peer grading are likely to
come from both an improved class design as well as statistical modeling.

Our results show that Models \PGthree (with coupled grader score and
reliability) and \PGtwo (with temporal coherence) yield the best results,
with Model \PGthree outperforming the other models with respect to most metrics.
But the single change that provides the
most significant gains in accuracy is obtained by estimating each grader's bias (Model \PGonebias). This simple
model is responsible for 95\% of our reduction in RMSE.
The other changes 
all contribute comparatively smaller improvements
to a more accurate model.

Our evaluation setup also allows us to test how accurate we would have been, had we had more than four grades per student. If the class had increased the number of grades that each student received to five (instead of four), our model could reduce RMSE error on the first and second offering of HCI to 4.19 and 4.36  respectively.

Surprisingly while modeling grader bias is particularly effective, modeling grader precision does little to improve our
performance. To dig deeper into this result we test our
model on a synthetic dataset --- one generated exactly from
Model \PGone . When using this synthetic data with only
four grades per student it is difficult for the model to
correctly estimate grader reliability. Modeling variance for
each grader only seems to have a notable impact when students grade many assignments 
(more than 10). This experiment also suggests why \PGthree is more useful than \PGone .
Though \PGone contains more expressive power
than \PGthree, estimating only two parameters for grader reliability ($\theta_0$ and $\theta_1$) is more statistically tractable with only
four grades per student than estimating a reliability, $\tau_v$, for
each grader. 
\vspace{-3mm}
\subsection{Fairness and efficiency in peer grading}
One of the advantages of using a probabilistic model for peer grading is
that we can obtain a belief distribution over grades (as opposed to a single score) for each student. These distributions give us a natural way of calculating how confident the model is when it predicts a grade for a student.
 The fact that the confidence results can be trusted
 open up the possibility of a more equitable allocation of graders. For example, at a given point midway through the peer grading process, our model may be highly confident
in its prediction for a given student's score, but very unsure in its prediction for another student. In this situation, to ensure that each student
gets fair access to quality feedback, we could reassign graders
to gradees such that submissions which have low-confidence
scores are given to more and/or better graders.

The first step towards more fair allocation of grades is to ask ourselves: how accurate are our
estimates of confidence?
For example, we would like to know how to interpret what it means in practice when our Bayesian
model is 90\% confident that its prediction of a learner's true score is within 10pp of the actual true score.

To better understand our confidence estimates, we run the following experiment: We first performed a large number of
peer grading simulations on ground truth. From each
simulation we calculate how confident our model is that
the grade it predict for the ground truth submission is
within 5\%, 7\%, and 10\%, of the true score, respectively. 
We then bin the estimated confidences into ranges 0-5\%, 5-10\%, etc.
After collecting over 5000 predictions per range, 
we test the pass rate of each range. For example, suppose we select four assessments of the same ground truth submission in a simulation. If our model reports a 72\% confidence --- based on those four assessments --- that our predicted grade is within 5pp
of the true score, we
add that estimate to the set of predictions in the 70\% to 75\%
confidence range. When we test this confidence range the example prediction ``passes'' if its estimate is in fact within 5pp of the ground truth score.

One worry is that our model might be overconfident about
its predictions even when wrong. However the results, shown in Figure~\ref{fig:confidence}, demonstrate
that our confidence estimates are on the conservative side ---
for example over 95\% of the time that our model claims it is between 90 and 95\% confident of a prediction, the model's estimate is correct. 

Since we have reason to believe that our confidence values are accurate, we can employ our posterior belief distributions to better allocate grades.  To understand how much benefit we could get out of improved grade allocation, we estimate at what point in the grading process we were confident about each submission's score. For each homework assignment, we simulate grading taking place in rounds. In the first round, we only include the first grade submitted by each grader (which may have been a ground truth grade). In the second round, we included the first two, etc. For each round we run our model using the corresponding subset of grades and count the number of submissions  for which we are over 90\% confident that our predicted grades were within 10pp of the student's true grade. 

After only two rounds of grading we are highly confident in our estimated grade for 15\% of submissions (this generally means that the submission has a grade close to the assignment mean, and has two similar grades from graders). Figure~\ref{fig:piechart} shows how the set of confident submissions grows over the grading rounds. Our experiment demonstrates a clear opportunity for grades to be reallocated as well as a pressing need for some submissions to get more grades. For 54\% of students, after all rounds, we are still unsure of their submission's true score.
\vspace{-4mm}
\subsection{Graders in the context of the MOOC}
\begin{figure*}[t!]
\begin{center}
\subfigure[]{
\raisebox{3mm}{
\includegraphics[width=.27\textwidth]{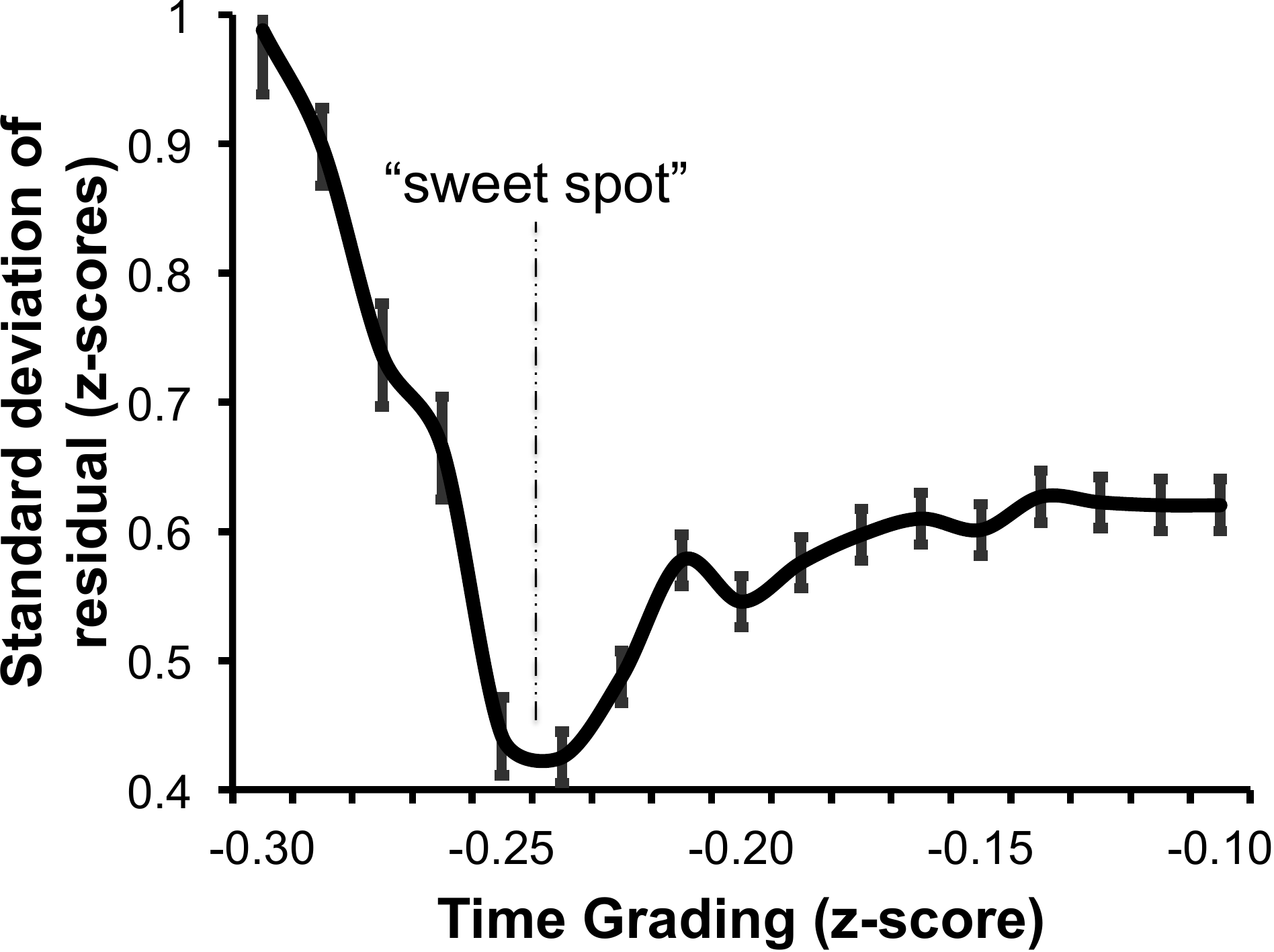}
}
\label{fig:time_v_residual}
}\;\;
\subfigure[]{
\includegraphics[width=.27\textwidth]{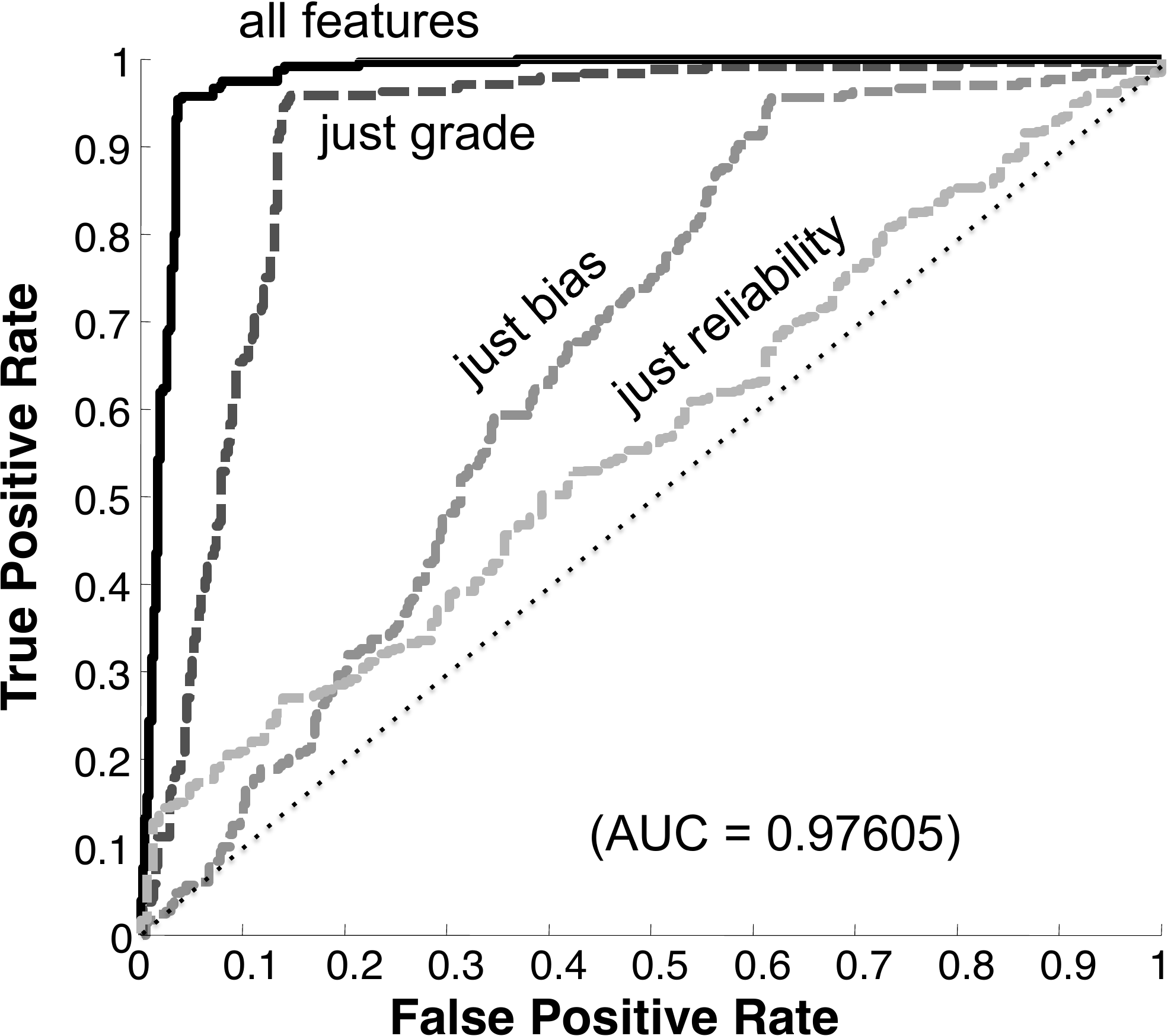}
\label{fig:roc}
}\;\;
\subfigure[]{
\includegraphics[width=.27\textwidth]{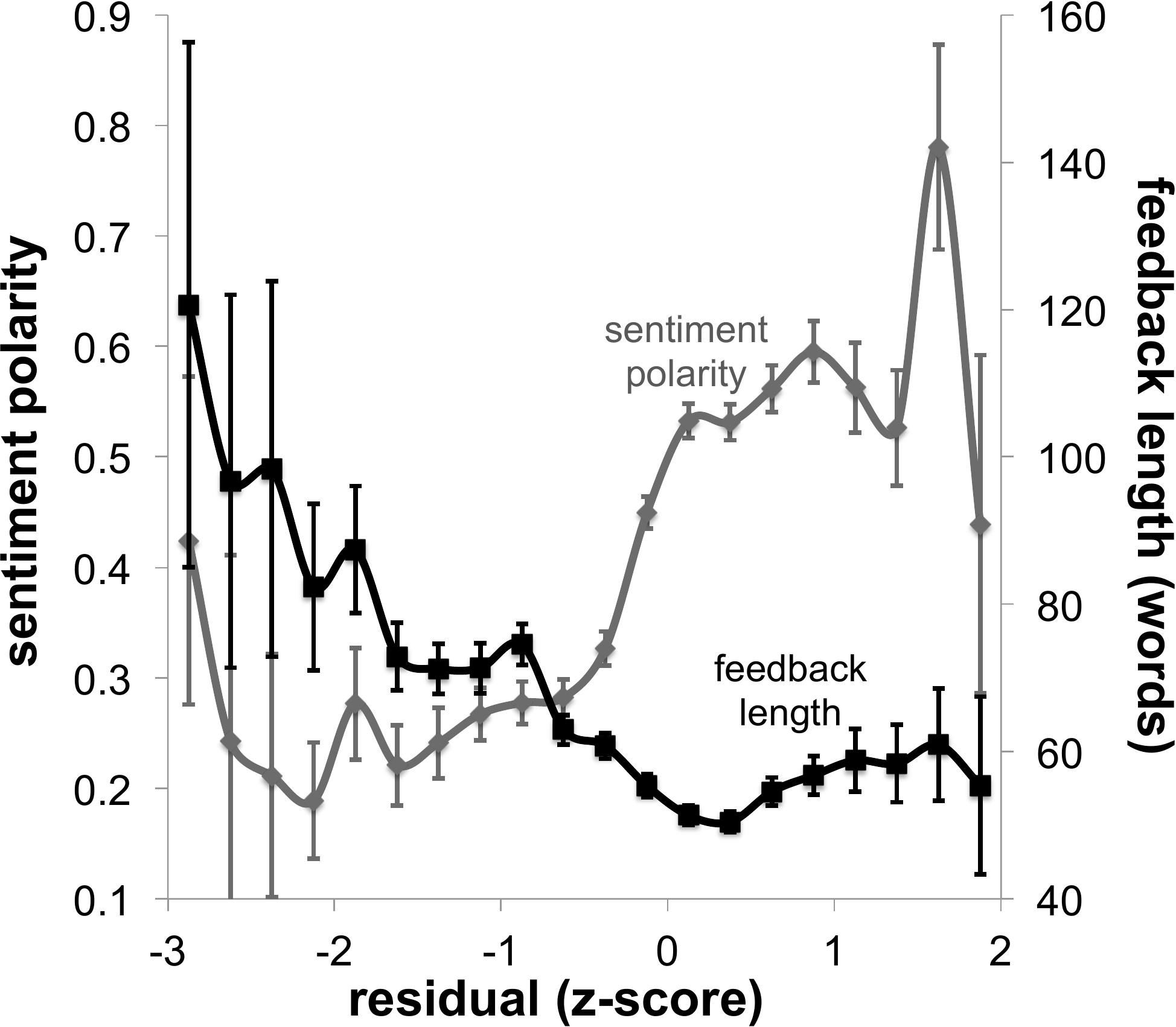}
\label{fig:comments}
}
\end{center}\vspace{-7mm}
\caption{\scriptsize
\subref{fig:time_v_residual}  Grader consistency (measured using standard deviation of grading residual) as
a function of time spent grading.
\subref{fig:roc} ROC curve comparing performance (with linear SVM) at predicting future class participation given
a student's grade, bias, reliability or all three.
\subref{fig:comments} Commenting style (length of comment and sentiment polarity) as a function of grading residual.
}\vspace{-4mm}
\end{figure*}
Applying probabilistic models to peer grading networks allows us to increase our grade accuracy and better allocate
what submissions students should grade. Another product
of our work is an assignment --- with a belief distribution --- for a true score, grader bias and grader reliability
for each student. We can use this large dataset to derive
new understanding about peer grading as both a formative
and summative assessment. We focus our investigation on
two questions, (1) what factors influence  how well a student
grades? and (2) how does grading ability affect future class
performance in a MOOC?
\vspace{-4mm}
\paragraph{Influential factors for grader ability} 
To explore what
factors influence how well a student grades we compare
grading residual (how far off a grader's score is from our
model estimated true score) to: time spent grading, grader grade, and gradee grade. 

Time spent grading shows a particularly interesting trend
(Figure~\ref{fig:time_v_residual}). As hypothesized, students that ``snap grade''
their peers' work (the students whose time spent grading has
a $z$-score of less than -0.30), are both unreliable (the
variance of their residuals is over 1 standard deviation
away from the gradee's true score) and tend to slightly
inflate grades. More surprising is that over the tens of
thousands of grades, there is a ``sweet spot'' of time spent
grading. Students who grade assessments with
a time that has a $z$-score of around -0.25 have significantly
lower residual standard deviations (with $p$-value < 0.001, diff = 0.3 standard
deviations) than students who take a long time to grade
(i.e., time spent grading has a $z$-score > -0.20). This sweet
spot is only visible when we look at normalized grading
times. For most assignments in the HCI class, the sweet
spot corresponds to around 20 minutes grading. This may reflect both that with any less time a grader does 
not have enough of a chance to fully examine her gradee's work, and that 
a long grading session may mean that the grader had trouble understanding 
some facet of the submission. 

Examining the relationship between grader grade, gradee
grade and how they affect the residual also shows a set of
notable trends. Graders that score higher on assignments
have close to monotonically decreasing biases (Figure~\ref{fig:gradergrade_v_residual}).
Getting a better grade on the homework in general makes
students more reliable graders; with the notable exception
that the students that get the best grades (+1.75 $z$-score)
are not as accurate as the students who do very well (+.75
$z$-score, $p$ = 0.04). The superlative submissions --- both the
best and the worst --- are the easiest to grade, and the submissions which are one standard deviation below the mean
are the hardest (Figure~\ref{fig:gradeegrade_v_residual}). Finally, our results show
that students are least biased when grading peers with
similar score (Figure~\ref{fig:gradergradee}). The best students significantly
downgrade the worst submissions and the worst students
notably inflate the best submissions.

In addition to numerical scores, graders were asked to provide feedback in the form of
free form text comments to their gradees. In order to understand the relationship between
grading performance and commenting style, we compare
grading residual against the comment length as well as sentiment polarity of the comment (Figure~\ref{fig:comments}).
To measure the polarity of a comment, we use the sentiment analysis
word list from~\cite{nielsen11} and implement a simple sentiment analyzer that returns a (normalized)
polarity score (positive or negative) 
proportional to the sum of word valences over the comment.
For both comment length and polarity, we filter out all non-English words.
We observe that comments that correspond to larger negative residuals are typically significantly longer,
suggesting perhaps that students write more about the weaknesses of a submission than strong points.
That being said, we observe that overall, the comments mostly range in polarity from neutral
to quite positive, suggesting that rather than being highly negative to some submissions,
many students make an effort to be balanced in their comments to peers.
\vspace{-4mm}
\paragraph{Grader ability and future performance}
We also tested what signal grading ability has with predicting future participation.
Based on the theory that the best graders are
intrinsically motivated, we hypothesized that being a reliable grader would add a different dimension of information
to a student's engagement which we should be able to use to
better predict future engagement. We tested this hypothesis
by constructing a classification task in which we predict
whether a student would participate in the next assignment
(or conversely which students would ``drop out''). In addition to the student's grade, we experimented with including
grader bias and reliability as features in a linear classifier.
Our results (Figure~\ref{fig:roc}) show that including grader bias and
reliability improved our predictive ability by 5pp from an
area under the curve (AUC) score of 0.93 to an AUC of 0.98.
Properties about how a student grades, captures a dimension of their engagement which is missed by their assignment grade.
\vspace{-3mm}
\section{Related work}\label{sec:relatedwork}
The statistical models we present in this paper can be seen as part of a long tradition of models which have been proposed for the purposes of aggregating information from noisy human labelers or workers.
Many of these works adapt classical item-response theory (IRT) models~\cite{baker01} to the problem of
``grading without an answer key'' and appear in the literature from educational aptitude testing~\cite{johnson96,rogers10,mislevy99}, to
cultural anthropology~\cite{batchelder88,karabatsos03}, and more recently to HCI in the context of human computation and crowdsourcing~\cite{whitehill09}.
In educational testing, for example, Johnson~\cite{johnson96} and Rogers et al.~\cite{rogers10}
propose models for combining human judgements of essays.  
These papers analyze \emph{dedicated human graders} who each evaluated hundreds
of essays, allowing for a rich model to be fitted on a per-grader basis.  In contrast, with peer
grading in MOOCs, each student only assesses a handful of assignments, necessitating more constrained models.



In a recent paper, and in a setting perhaps most similar to our
own, Goldin et al.~\cite{goldin11,ashley11,goldin12} use Bayesian models for peer grading in a smaller scale classroom setting. As in our
own work,~\cite{goldin12} posits a grader bias, and in fact incorporates
rubric-specific biases, but does not consider many of the
issues raised here such as grading task reallocation or the
relationship between grader bias and student engagement,
for example.

One of the central themes of the crowdsourcing literature, that of balancing label accuracy against labor cost, is one which MOOC peer grading 
systems must contend with as well.
In such problems, one typically receives a number of noisy labels (for example in an image tagging task)
and the challenge lies in (1) resolving the ``correct'' label (often discrete, but sometimes continuous) 
and (2) deciding whether to hire more labelers for a given task.  
Explosion of interest in recent years has led to widespread applications of crowdsourcing~\cite{bachrach12,kamar12}.
For example in image annotation, Whitehill et al.~\cite{whitehill09} present a method similar to our
own in which they model discrete ``true image labels'' as well as labeler accuracy.
While our work draws from the crowdsourcing literature, the problem of peer grading is unique in several ways. For example,
the fact that the graders are also gradees in peer grading is
quite different from typical crowdsourcing settings in which there is a dichotomy between the labelers
and the items being labeled, and motivates different models (such as Model \PGthree). In crowdsourcing applications,
 the end goal often lies in determining the true labels rather than to understand anything
about the labelers themselves, whereas in peer grading, as we have shown, the insights that we can glean about the
graders have educational value.

A similar problem to peer-grading is the paper assignment problem for the peer review process in academic conferences.  
While related in that the central challenge of both problems involves fusing disparate human opinions about open-ended creative work,
many of the specific challenges are distinct.
For one, side information plays a much larger role in peer review, where conference chairs typically
rely heavily on personal or elicited knowledge of reviewer expertise or citation link structure to assign reviewer roles~\cite{charlin11}.
Peer grading on the other hand seems less sensitive to personal preferences, where a single submission should be equally
well graded by a large fraction of students in the course.

\vspace{-2mm}
\section{Discussion and Future work}
Our paper presents methods for making large scale peer
grading systems more dependable, accurate, and efficient.
In particular, we show that there is much to be gained
by maintaining estimates of grader specific quantities such
as bias and reliability. In addition to improving peer grading
accuracy by up to 30\%, these quantities give a unique insight
into peer grading as a formative and summative assessment.

There remain a number of issues to be addressed in future
work. We have considered the problem of determining which
submissions need to be allocated additional graders. However, deciding which grader is best for evaluating a particular
submission is an open problem whose solution could depend on
a number of variables, from the writing styles of the grader
and gradee to their respective cultural or linguistic backgrounds, a particularly important issue for the global scale
course rosters that arise in MOOCS.

Another issue arises from our study of the biases from graders
who do not spend adequate time on grading. Incentivizing these students to provide careful and high quality
feedback to their peers is a question of paramount importance for open-access courses. Using model \PGthree for scoring, as we discussed, makes
a student's score dependent on grading performance, and
may be one way to build a justified, incentive directly into the scoring mechanism. Understanding this and other scoring rules
from a game theoretical perspective remains for future work. 

Finally, it is not clear how to present scores which are calculated by a complicated peer grading model to a students. While this communication
might be easy when a student's final grade is simply set
to be the mean or median of peer grades, does each student need to know the inner workings of
a more sophisticated statistical backend? Students may be unhappy with the lack of transparency in
grading mechanisms, or on the other hand might feel more satisfied with their overall grade.

As MOOCs become more widespread, the need for reliable
grading and feedback for open ended assignments becomes
ever more critical. The most scalable solution that has been
shown to be effective is peer grading. By addressing the
shortcomings of current peer grading systems, we hope that
students everywhere can get more from peer grading and
consequently, more from their free online, open access educational experience.

\vspace{-3mm}
\section*{Acknowledgments}

{\footnotesize
We thank Chinmay Kulkarni and Scott Klemmer for providing assistance with the HCI datasets and Leonidas Guibas and John Mitchell
for discussions and support. Jonathan Huang is supported by an NSF 
CI Fellowship.
}

%
{\footnotesize

\begin{thebibliography}{10}

\bibitem{ashley11}
K.~Ashley and I.~Goldin.
\newblock Toward ai-enhanced computer-supported peer review in legal education.
\newblock In {\em 24th International Conference on Legal Knowledge and
  Information Systems (JURIX)}, volume 235, 2011.

\bibitem{bachrach12}
Y.~Bachrach, T.~Minka, J.~Guiver, and T.~Graepel.
\newblock {How To Grade a Test Without Knowing the Answers - A Bayesian
  Graphical Model for Adaptive Crowdsourcing and Aptitude Testing}.
\newblock In {\em The 29th Annual International Conference on Machine
  Learning}, ICML '12, 2012.

\bibitem{baker01}
F.~Baker.
\newblock {\em The basics of item response theory}.
\newblock ERIC Clearinghouse on Assessment and Evaluation, University of
  Maryland, College Park, 2001.

\bibitem{batchelder88}
W.~H. Batchelder and A.~K. Romney.
\newblock Test theory without an answer key.
\newblock {\em Psychometrika}, 53:71--92, 1988.

\bibitem{charlin11}
L.~Charlin, R.~Zemel, and C.~Boutilier.
\newblock A framework for optimizing paper matching.
\newblock In {\em Proceedings of Uncertainty in Artificial Intelligence
  (UAI'11)}, 2011.

\bibitem{geman84}
S.~Geman and D.~Geman.
\newblock Stochastic relaxation, gibbs distributions, and the bayesian
  restoration of images.
\newblock {\em IEEE PAMI}, (6):721--741, 1984.

\bibitem{goldin12}
I.~Goldin.
\newblock Accounting for peer reviewer bias with bayesian models.
\newblock In {\em Proceedings of the Workshop on Intelligent Support for
  Learning Groups at the 11th International Conference on Intelligent Tutoring
  Systems}, 2012.

\bibitem{goldin11}
I.~M. Goldin and K.~D. Ashley.
\newblock Peering inside peer review with bayesian models.
\newblock In {\em Proceedings of the 15th international conference on
  Artificial intelligence in education}, AIED'11, pages 90--97, Berlin,
  Heidelberg, 2011. Springer-Verlag.

\bibitem{johnson96}
V.~E. Johnson.
\newblock On bayesian analysis of multi-rater ordinal data: An application to
  automated essay grading.
\newblock {\em Journal of the American Statistical Association}, 91:42--51,
  1996.

\bibitem{kamar12}
E.~Kamar, S.~Hacker, and E.~Horvitz.
\newblock Combining human and machine intelligence in large-scale
  crowdsourcing.
\newblock In {\em In AAMAS}, 2012.

\bibitem{karabatsos03}
G.~Karabatsos and W.~Batchelder.
\newblock Markov chain estimation for test theory without an answer key.
\newblock {\em Psychometrika}, 68(3):373--389, 2003.

\bibitem{kulkarni13}
C.~Kulkarni, K.~Pang-Wei, H.~Le, D.~Chia, K.~Papadopoulos, D.~Koller, and S.~R.
  Klemmer.
\newblock Scaling self and peer assessment to the global design classroom.
\newblock In {\em Proceedings of CHI'13 (to appear)}, 2013.

\bibitem{mislevy99}
R.~J. Mislevy, R.~G. Almond, D.~Yan, and L.~S. Steinberg.
\newblock Bayes nets in educational assessment: Where the numbers come from.
\newblock In {\em Proceedings of the fifteenth conference on uncertainty in
  artificial intelligence}, pages 437--446. Morgan Kaufmann Publishers Inc.,
  1999.

\bibitem{nielsen11}
F.~{\AA}. Nielsen.
\newblock A new anew: Evaluation of a word list for sentiment analysis in
  microblogs.
\newblock {\em arXiv preprint arXiv:1103.2903}, 2011.

\bibitem{rogers10}
S.~Rogers, M.~Girolami, and T.~Polajnar.
\newblock Semi-parametric analysis of multi-rater data.
\newblock {\em Statistics and Computing}, 20(3):317--334, July 2010.

\bibitem{russell04}
A.~A. Russell.
\newblock Calibrated peer review - a writing and critical-thinking
  instructional tool.
\newblock {\em Teaching Tips: Innovations in Undergraduate Science
  Instruction}, page~54, 2004.

\bibitem{sadler06}
P.~M. Sadler and E.~Good.
\newblock The impact of self-and peer-grading on student learning.
\newblock {\em Educational assessment}, 11(1):1--31, 2006.

\bibitem{whitehill09}
J.~Whitehill, P.~Ruvolo, T.~fan Wu, J.~Bergsma, and J.~Movellan.
\newblock Whose vote should count more: Optimal integration of labels from
  labelers of unknown expertise.
\newblock In {\em Advances in Neural Information Processing Systems 22}, pages
  2035--2043. MIT Press, 2009.

\end{thebibliography}

}
%
%


\end{document}